%% file: main.tex
\documentclass{article}

\usepackage{microtype}
\usepackage{graphicx}
\usepackage{subfig}
\usepackage{booktabs} % for professional tables

\usepackage{multirow}
\usepackage{hyperref}

\usepackage{threeparttable}

\usepackage[accepted]{icml2025}
\input{pre}

\usepackage{amsmath}
\usepackage{amssymb}
\usepackage{mathtools}
\usepackage{amsthm}

\usepackage[capitalize,noabbrev]{cleveref}

\theoremstyle{plain}

\theoremstyle{definition}

\theoremstyle{remark}

\usepackage[disable,textsize=tiny]{todonotes}
\icmltitlerunning{Validating Deep Models for Alzheimer’s $^{18}$F-FDG PET Diagnosis Across Populations: A Study with Latin American Data}

\begin{document}

\twocolumn[
\icmltitle{Validating Deep Models for Alzheimer’s $^{18}$F-FDG PET Diagnosis Across Populations: A Study with Latin American Data}



\icmlsetsymbol{equal}{*}

\begin{icmlauthorlist}
\icmlauthor{Hugo Massaroli}{ditella,uba}
\icmlauthor{Hernan Chaves}{fleni}
\icmlauthor{Pilar Anania}{fleni}
\icmlauthor{Mauricio Farez}{fleni}
\icmlauthor{Emmanuel Iarussi}{ditella}
\icmlauthor{Viviana Siless}{ditella}
\end{icmlauthorlist}

\icmlcorrespondingauthor{Viviana Siless}{viviana.siless@utdt.edu}

\icmlaffiliation{ditella}{Business School, Di Tella University, Argentina}
\icmlaffiliation{fleni}{Fleni Institute, Buenos Aires, Argentina}
\icmlaffiliation{uba}{Departamento de Computación, Universidad de Buenos Aires, Buenos Aires, Argentina}

\icmlkeywords{Machine Learning, ICML}

\vskip 0.3in
]



\printAffiliationsAndNotice{}  % leave blank if no need to mention equal contribution

\begin{abstract}
Deep learning models have shown strong performance in diagnosing Alzheimer’s disease (AD) using neuroimaging data, particularly $^{18}$F-FDG PET scans, with training datasets largely composed of North American cohorts such as those in the Alzheimer’s Disease Neuroimaging Initiative (ADNI). 
However, their generalization to underrepresented populations remains underexplored. 
In this study, we benchmark convolutional and Transformer-based models on the ADNI dataset and assess their generalization performance on a novel Latin American clinical cohort from the FLENI Institute in Buenos Aires, Argentina.
We show that while all models achieve high AUCs on ADNI  (up to .96, .97), their performance drops substantially on FLENI (down to .82, .80, respectively), revealing a significant domain shift. 
The tested architectures demonstrated similar performance, calling into question the supposed advantages of transformers for this specific task.
Through ablation studies, we identify per-image normalization and a correct sampling selection as key factors for generalization. 
Occlusion sensitivity analysis further reveals that models trained on ADNI, generally attend to canonical hypometabolic regions for the AD class, but focus becomes unclear for the other classes and for FLENI scans. 
These findings highlight the need for population-aware validation of diagnostic AI models and motivate future work on domain adaptation and cohort diversification.
\end{abstract}

\input{secs/intro}
\input{secs/relatedwork}
\input{secs/data}
\input{secs/methods}

\input{secs/results}

\input{secs/conclusions}


\bibliography{main}
\bibliographystyle{icml2025}
\end{document}

%% file: pre.tex
\usepackage{dsfont}
\usepackage{amsthm}
\usepackage{overpic}
\usepackage{contour}
\contourlength{1pt}
\usepackage{xspace}
\usepackage{sidecap}
\usepackage{rotating}
\usepackage[normalem]{ulem}
\usepackage{amsmath}
\usepackage{xcolor,colortbl}
\usepackage{tabularx}
\usepackage{soul}
\usepackage{url}
\usepackage{amsfonts}
\usepackage{comment}
\usepackage{booktabs}
\usepackage{etoolbox}
\usepackage{cancel}

\newlength{\indentlaenge}

%% file: secs/intro.tex
\section{Introduction}

Alzheimer’s disease (AD) is the most common cause of dementia, accounting for an estimated 60–70\% of the 57 million dementia cases worldwide as of 2021~\cite{WHO2021dementia}. 
With aging populations, this number is projected to rise sharply in the coming decades.
Early and accurate diagnosis is essential for managing disease progression and informing treatment strategies. 
$^{18}$F-fluorodeoxyglucose positron emission tomography ($^{18}$F-FDG PET) has proven to be a reliable biomarker of cerebral glucose metabolism, revealing characteristic patterns of hypometabolism associated with AD \cite{chetelat2020amyloid}.

Recent years have witnessed the application of deep learning methods to neuroimaging for automated AD diagnosis \cite{ding2019deep}. 
Models trained on large datasets such as the Alzheimer’s Disease Neuroimaging Initiative (ADNI) have achieved high classification accuracy. However, there is growing concern that such models may not generalize well to demographically or clinically distinct populations \cite{Arora2023,Yang2024}.

In this work, we explore this challenge by comparing the performance of three deep learning models—a convolutional neural network (CNN), a transformer-based architecture, and a lightweight ResNet \cite{resnet} variant—trained on ADNI and evaluated on a novel Latin American dataset from the Fundación para la Lucha contra las Enfermedades Neurológicas de la Infancia (FLENI) Institute in Argentina. 
This cross-cohort evaluation enables a realistic assessment of model robustness in diagnostic AI applications.

Our contributions are threefold:
\begin{itemize}
    \item We benchmark and compare three architectures under matched training settings, reporting both in-distribution and out-of-distribution performance.
    \item We conduct ablation studies on preprocessing strategies and input representations to identify key factors impacting generalization.
    \item We provide visual diagnostic analyses to interpret model decisions across domains, revealing differences in spatial attention patterns.
\end{itemize}

Our findings expose consistent performance degradation on the FLENI dataset and support the claim that standard benchmarks alone are insufficient for evaluating clinically deployable models.

%% file: secs/relatedwork.tex
\section{Related Work}
Research on automated Alzheimer's disease diagnosis using neuroimaging has been active for over a few decades. Early approaches focused on handcrafted features derived from structural MRI and FDG-PET, including hippocampal atrophy and hypometabolic patterns in temporoparietal regions—both established biomarkers of AD progression~\cite{frisoni2010clinical, weiner2015impact}.

More recently, deep learning methods have achieved state-of-the-art performance by automatically learning representations from raw imaging data. 
In particular, Convolutional Neural Networks (CNNs) have been the dominant architecture in AD neuroimaging tasks, particularly for image classification and disease stage prediction~\cite{ding2019deep}. 
These models are well-suited for capturing local spatial patterns and have demonstrated robustness in both binary and multi-class AD classification problems.

More recently, Transformer-based architectures~\cite{vaswani2017attention}, originally developed for natural language processing, have recently gained traction in medical imaging due to their ability to model long-range dependencies via self-attention mechanisms~\cite{shamshad2023transformers,he2023transformers,li2023transforming}. 
Unlike CNNs, which focus on localized receptive fields, transformers can capture global contextual relationships, which is particularly valuable in neuroimaging where subtle and spatially distributed patterns may be predictive of disease. 
However, to the best of our knowledge, Transformer-based models have not yet been introduced into FDG-PET analysis for Alzheimer's disease.

Whether models trained on one population can generalize effectively to diverse clinical cohorts remains an open and critical question in the field ~\cite{Arora2023,Yang2024}.
Most AD classification studies are conducted on homogeneous datasets such as ADNI, which does not reflect the diversity of real-world populations. Evaluating models on independent cohorts, particularly from underrepresented regions like Latin America, is essential for assessing their true clinical utility.

Despite growing interest in fair and generalizable AI systems, to our knowledge, most studies in AD diagnosis continue to rely on demographically homogeneous datasets. This poses significant limitations in regions where population health characteristics, imaging protocols, and healthcare infrastructure differ from those represented in benchmark datasets. To our knowledge, this is the first study to explicitly evaluate the generalization of both CNN and transformer-based architectures trained on ADNI to a Latin American cohort using FDG-PET data.

%% file: secs/data.tex
\section{Data}

We used multiple FDG-PET datasets to assess model performance and generalization across heterogeneous cohorts: ADNI, FLENI100, and FLENI600.

\subsection{ADNI Dataset}
The Alzheimer’s Disease Neuroimaging Initiative (ADNI) is a longitudinal, multicenter study aimed at developing clinical, imaging, genetic, and biochemical biomarkers for the early detection of Alzheimer’s disease (AD). We included subjects from ADNI1, ADNI-GO, ADNI2, and ADNI3, focusing on FDG-PET scans.

The dataset comprises 3,762 pre-processed image records from 1,687 unique subjects. Each subject contributed multiple scans due to longitudinal follow-up. The image volumes are composed of 96 DICOM slices each, representing 3D brain scans of dimensions $160 \times 160 \times 96$ voxels with an isotropic resolution of 1.5 mm.

ADNI diagnostic labels are assigned clinically at each visit and span five classes: cognitively normal (CN), subjective memory concerns (SMC), early mild cognitive impairment (EMCI), mild cognitive impairment (MCI), late mild cognitive impairment (LMCI), and Alzheimer’s dementia (AD). For this study, we excluded SMC and merged EMCI, LMCI, and MCI into a unified MCI category.
To facilitate comparison with FLENI’s biomarker-driven labeling, we created two label variants:
\begin{itemize}
    \item \texttt{visit953}: Labels are based on the clinical diagnosis closest to the FDG-PET scan, with two classes:
    \begin{itemize}
        \item \textbf{AD}: Subjects diagnosed with Alzheimer’s disease (AD) at scan time, or with mild cognitive impairment (MCI) who later converted to AD.
        \item \textbf{Non-AD}: Cognitively normal (CN) subjects and MCI cases that did not convert to AD, grouped to match FLENI’s non-AD biomarker profile.
    \end{itemize}
    Each subject contributes one scan. For MCI-to-AD converters, we used the latest scan labeled AD; for CN-to-MCI converters, the earliest scan labeled CN. MCI subjects who neither converted nor originated from CN were excluded.
    
    \item \texttt{last}: Labels reflect the diagnosis at the subject’s final recorded visit, with three classes: CN, MCI, and AD.
\end{itemize}

\begin{table*}[t]
\centering
\caption{Demographics and class distribution across datasets.}
\label{tab:demographics}
\begin{tabularx}{\textwidth}{*{8}{X}}
\toprule
Dataset & Labeling &Gender & AD & MCI & CN & Non-AD \tnote{*} & Total \\
\midrule
ADNI & last & Male   & 299 & 428 & 204 & --- & 931 \\
                         &      & Female & 206 & 306 & 231 & --- & 743 \\
ADNI & visit953 & Male   & 291 & --- & --- & 229 & 520 \\
                    &     & Female & 195 & --- & --- & 238 & 433 \\
FLENI & FLENI100 & Male   & 33  & --- & --- & 24  & 57 \\
           &    & Female & 27  & --- & --- & 16  & 43 \\
FLENI & FLENI600 & Male   & 147 & --- & --- & 154 & 301 \\
           &    & Female & 173 & --- & --- & 120 & 293 \\
\bottomrule
\end{tabularx}
\end{table*}

\subsection{FLENI Datasets}
The FLENI datasets originate from the Fundación para la Lucha contra las Enfermedades Neurológicas de la Infancia (FLENI) in Argentina. These datasets include patients evaluated with both PIB and FDG PET tracers. Image acquisition was performed using a General Electric Discovery 690 PET/CT scanner.

Each subject underwent a two-step scanning protocol. First, 13.44 mCi of \textsuperscript{11}C-PIB was administered intravenously, followed by a 50-minute uptake period and a 20-minute scan. Thirty minutes later, 6.49 mCi of \textsuperscript{18}FDG was administered, followed by a 30-minute rest and a subsequent 20-minute scan. Each FDG study consists of 47 DICOM slices forming a 3D image of size $128 \times 128 \times 47$ with voxel dimensions of $2 \times 2 \times 3.27$ mm (non-isotropic). When using the Transformer model with input samples containing 77 axial slices, the data is resized to $128 \times 128 \times 77$ voxels using nearest-neighbor interpolation.

\textbf{FLENI100:} This subset consists of 100 studies with biomarker-based diagnoses. AD status is determined by positive cerebrospinal fluid (CSF) beta-amyloid and phosphorylated tau levels or corresponding PIB PET positivity, enabling early detection before dementia symptoms appear. Control subjects were biomarker-negative. Both AD and Non-AD cohorts may include MCI patients.

\textbf{FLENI600:} This larger dataset includes 594 studies labeled based on PET imaging alone. AD subjects exhibit characteristic cortical FDG hypometabolism and PIB retention. Controls present with normal FDG uptake and no PIB signal. Age information is not available for FLENI600. As in FLENI100, both cohorts may include MCI subjects.

\subsection{Cohort Demographics and Label Distribution}
Table~\ref{tab:demographics} summarizes the gender and diagnostic breakdown for each dataset and label strategy.
Age statistics for ADNI and FLENI100 are available, with ADNI subjects ranging from 55 to 96 years and FLENI100 from 43 to 90. FLENI600 age data is not recorded.

%% file: secs/methods.tex
\section{Methods}

Preprocessing included skull stripping and resampling to a uniform voxel grid of $128 \times 128 \times 77$, corresponding to the sagittal, coronal, and axial planes, respectively.
Intensity values were normalized using z-score normalization either globally or per subject. For CNN-based models, we extracted 16 equidistant axial slices. For Transformers, we used full 3D volumes with orthogonal plane projections.

\subsection{Model Architectures}
We implemented and evaluated three neural network architectures:
\begin{itemize}
    \item \textbf{Inception CNN}: A 2D convolutional model based on InceptionV3 operating on 16 axial FDG-PET slices arranged into a $4 \times 4$ grid, as in~\cite{ding2019deep}. It uses ReLU activations, dropout, and softmax output.
    \item \textbf{Transformer}: Adapted from Medical Transformer \cite{Jun2024}, it integrates a ResNet-18 encoder and multi-head self-attention across anatomical planes with positional encodings. Each brain slice is represented by a single token with 64 features. Tokens from all three anatomical planes are concatenated, resulting in a sequence of 333 tokens per image. The transformer layer employs 4 attention heads to contextualize the token sequence. Following this, the architecture includes a dropout layer, two fully connected layers, and a softmax output. To ensure a fair comparison with the other models, this architecture can also be configured to use 16 axial slices instead of 77.
   \item \textbf{Pruned ResNet  (P-ResNet)}: A streamlined model derived from the Transformer, using 16 axial slices and omitting the transformer encoder and sagittal/coronal tokens. Each slice is embedded to a single scalar (reduced from 64), resulting in a 16-dimensional input to the final fully connected layer. This architecture has 718,000 parameters, making it significantly lighter than the Inception and Transformer models.
\end{itemize}

All models were implemented in PyTorch 2.0.1 and trained using the Adam optimizer. Data augmentation included rotation, flipping, intensity jittering, and Gaussian noise.

\subsection{Training and Validation Protocol}
Models were trained on ADNI using 10-fold cross-validation with stratified splits. The best model was selected based on validation AUC-ROC. For generalization assessment, models were evaluated on the FLENI datasets without fine-tuning.

\begin{figure}[h]
    \centering

    \subfloat[General diagram]{\includegraphics[width=0.2\textwidth]{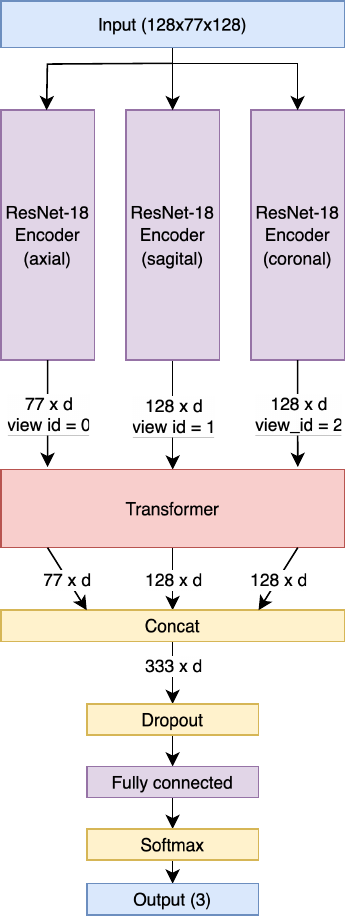}\label{fig:transformer_architecture_general}}
    \hfill
    \subfloat[Transformer block]{\includegraphics[width=0.2\textwidth]{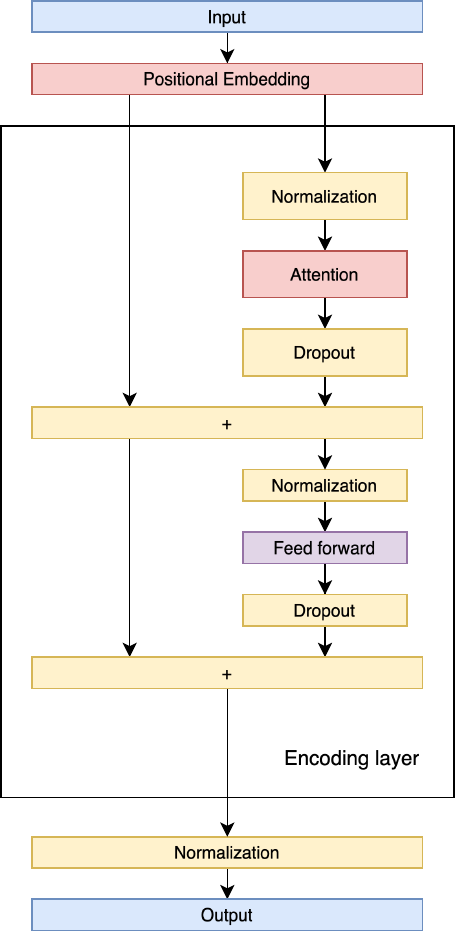}\label{fig:transformer_block}}

    \caption{Schematic of the Transformer architecture. Multi-view ResNet encoders process axial, sagittal, and coronal slices. Features are flattened into patches with positional encoding and passed to a transformer encoder.}
     \label{fig:transformer_architecture}
\end{figure}

Hyperparameters were selected via grid search. The Inception model used a learning rate of 0.0001 and a batch size of 32. Transformers and P-ResNet used a learning rate of 0.0005 and a batch size of 8. Cross-entropy loss was weighted by inverse class frequency. Early stopping was applied with a patience of 30 epochs. This is summarized in Table \ref{tab:hyperparams}.

Experiments were conducted on a computing server equipped with two NVIDIA A100 80GB GPUs and 256 GB of RAM.

\begin{table}[t]
\centering
\caption{Model hyperparameters and configurations.}
\label{tab:hyperparams}
\resizebox{\columnwidth}{!}{%
\begin{tabular}{lccccc}
\toprule
Model & LR & Batch Size & Dropout & Epochs & Patience \\
\midrule
Inception & 0.0001 & 32 & 0.6 & 100 & 30 \\
Transformer & 0.0005 & 8 & 0.4 & 100 & 30 \\
P-ResNet  & 0.0005 & 8 & 0.4 & 100 & 30 \\
\bottomrule
\end{tabular}
}
\end{table}

\subsection{Evaluation Metrics}
We report the area under the ROC curve (AUC), accuracy, sensitivity, and specificity for both ADNI (in-distribution) and FLENI (out-of-distribution) evaluations. Occlusion sensitivity maps were used for interpretability.

%% file: secs/results.tex
\section{Results}

\subsection{Model Performance}
We report all experimental configurations for Inception, Transformer and P-ResNet models in Table~\ref{tab:merged-results}, including normalization strategies, slice count, and AUC values across datasets.

In the context of sample selection strategies, \textbf{first} refers to using only the first scan per subject in the validation and test sets, ensuring that no individual contributes multiple data points during evaluation. \textbf{First w/train} extends this constraint to the training set as well, using only the first scan per subject throughout all data partitions. This strategy aims to evaluate model generalization in a more realistic clinical scenario where repeated measures are not available.
\begin{table*}[t]
\centering
\caption{Merged AUC ROC Results for Inception, Transformer and P-ResNet Models. Sample selection strategies: \textbf{first} uses the first scan per subject for validation and testing; \textbf{first w/train} applies the same restriction to training as well.}
\label{tab:merged-results}
\begin{tabular}{lccccccl}
\toprule
Model & Labeling & Classes & Slices & Normalization & ADNI Test & FLENI100 & FLENI600 \\
\midrule
\multirow{8}{*}{Inception}
& last & 3 & 16 & z-score global (first w/train) & 0.83 (0.02) & 0.63 (0.03) & 0.72 (0.03) \\
& last & 3 & 16 & z-score (first) & 0.82 (0.02) & 0.69 (0.09) & 0.78 (0.04) \\
& last & 3 & 16 & z-score (first w/train) & 0.83 (0.02) & \textbf{0.71} (0.05) & \textbf{0.81} (0.03) \\
& last & 3 & 16 & min-max (first w/train) & 0.82 (0.02) & 0.70 (0.03) & 0.80 (0.02) \\
& visit953 & 2 & 16 & z-score global & 0.95 (0.01) & 0.64 (0.04) & 0.68 (0.03) \\
& visit953 & 2 & 16 & z-score per-image & 0.96 (0.01) & \textbf{0.73} (0.04) & \textbf{0.82} (0.02) \\
& visit953 & 2 & 16 & min-max & 0.96 (0.01) & \textbf{0.73} (0.03) & 0.81 (0.02) \\
\midrule
\multirow{8}{*}{Transformer}
& last & 3 & 77 & min-max & 0.84 (0.02) & 0.70 (0.03) & 0.78 (0.03) \\
& last & 3 & 16 & min-max & 0.82 (0.01) & 0.68 (0.04) & 0.78 (0.01) \\
& visit953 & 2 & 77 & min-max & 0.95 (0.01) & \textbf{0.76} (0.02) & \textbf{0.81} (0.01) \\
& visit953 & 2 & 16 & min-max & 0.96 (0.01) & 0.72 (0.03) & \textbf{0.81} (0.02) \\
& last & 3 & 77 & z-score per-image & 0.84 (0.02) & 0.71 (0.03) & 0.79 (0.02) \\
& last & 3 & 16 & z-score per-image & 0.81 (0.01) & 0.66 (0.04) & 0.78 (0.02) \\
& visit953 & 2 & 77 & z-score per-image & 0.95 (0.01) & \textbf{0.74} (0.02) & \textbf{0.81} (0.02) \\
& visit953 & 2 & 16 & z-score per-image & 0.95 (0.01) & 0.71 (0.02) & 0.79 (0.02) \\
\midrule
\multirow{6}{*}{P-ResNet}
& last & 3 & 16 & min-max & 0.84 (0.01) & \textbf{0.74} (0.02) & \textbf{0.81} (0.01) \\
& visit953 & 2 & 16 & min-max & 0.96 (0.01) & 0.72 (0.03) & 0.80 (0.01) \\
& last & 3 & 16 & z-score per-image & 0.84 (0.01) & \textbf{0.74} (0.03) & \textbf{0.82} (0.01) \\
& visit953 & 2 & 16 & z-score per-image & 0.97 (0.00) & 0.71 (0.02) & 0.80 (0.01) \\
\bottomrule
\end{tabular}
\end{table*}

This comprehensive table merges all configurations and helps illustrate the trade-offs across normalization, model type, input strategy, and labeling. Best AUCs on FLENI datasets are highlighted in bold. A summary of top-performing settings remains in Table~\ref{tab:merged-results}.

\subsection{Stability and Convergence}

Training behavior varied across architectures. The Transformer model, when using 16 slices, exhibited less variation across folds and showed signs of overfitting, converging in an average of 19 epochs.
It was followed by the Inception model, with an average best epoch of 24.
In contrast, the Transformer (77 slices) and ResNet architectures showed fewer signs of overfitting, with an average convergence epoch of 30 and 46 epochs each. 

\subsection{Ablation Studies}

To isolate the impact of preprocessing and input configuration, we conducted a series of ablation experiments on every model. Table~\ref{tab:merged-results} summarizes how modifications such as slice selection strategy, image selection strategy, and normalization approach influenced generalization.

The results emphasize the importance of both spatial context and normalization scheme. 
Full axial coverage and per-image z-score normalization yielded the most robust cross-domain performance. 
In the Transformer model, reducing spatial input weakened generalization, especially on the FLENI dataset.

\subsection{Visual Diagnostics}

To further understand model behavior and domain-specific biases, we employed occlusion sensitivity analysis on the Inception model. This method involves masking out spatial patches of the input and quantifying the change in prediction confidence, thereby revealing regions critical for classification.

\begin{figure*}[t]
    \centering
    \includegraphics[width=1.0\textwidth]{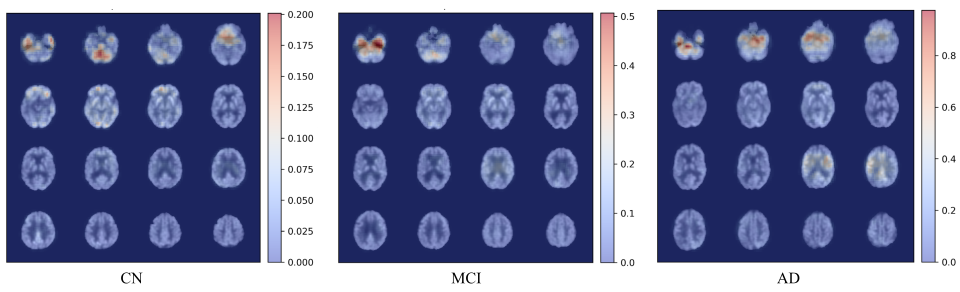} \\
    \vspace{0.3cm}
    \includegraphics[width=0.65\textwidth]{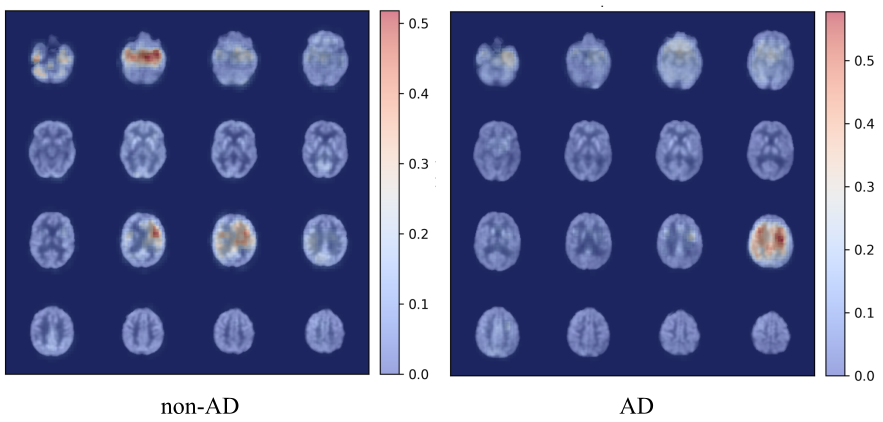}
    \caption{Occlusion sensitivity maps from the Inception model. Top: ADNI samples (CN, MCI, AD). Bottom: FLENI 600 samples (non-AD, AD). The highlighted regions indicate areas with the greatest impact on model predictions.}
    \label{fig:gradcam}
\end{figure*}

Figure~\ref{fig:gradcam} displays  occlusion maps from representative samples. In ADNI, we highlight three diagnostic classes (CN, MCI, AD), whereas FLENI 600 reflects binary classification (Non-AD vs. AD).

In ADNI, occlusion-based relevance scores tend to be stronger and more concentrated, suggesting more confident and localized decision patterns. Surprisingly, the regions that the network attends to in FLENI differ from those in ADNI, with more diffuse or inconsistent activation. This indicates possible structural or intensity mismatches between cohorts that may challenge the model’s learned representations. These qualitative findings mirror the observed quantitative performance drop and reinforce the clinical importance of validating models across diverse populations.

%% file: secs/conclusions.tex
\section{Discussion}

Our experiments show that while deep learning models for Alzheimer’s disease classification can achieve strong performance within a controlled training domain (e.g., ADNI), their generalization to external, demographically distinct cohorts remains a major challenge. All three models—Inception CNN, Transformer, and pruned ResNet—showed reduced performance on the FLENI dataset, with AUCs dropping by over 10 percentage points in some cases. This domain shift underscores the need to assess model robustness not just on held-out data from the same source, but across clinically realistic, geographically diverse populations.

All evaluated architectures—Inception, Transformer, and ResNet—achieved strong performance on both the in-distribution (ADNI) dataset and the out-of-distribution (FLENI) dataset. The best Transformer configuration reached an AUC of 0.95 on ADNI and 0.76/0.81 on FLENI100/FLENI600. P-ResNet followed closely with 0.84/0.74/0.82, showing strong cross-cohort generalization despite its compact architecture. Inception achieved comparable performance, notably with a best result of 0.96 on ADNI and 0.73/0.82 on FLENI. While no single model outperformed all others in every setting, Inception offered the strongest performance on ADNI without compromising generalization. These results highlight that effective generalization across cohorts does not necessarily require the most complex model, and underscore the importance of systematic cross-dataset validation.

Ablation studies further revealed that full-slice input and per-image normalization are key for generalization. Simplifying spatial input or altering preprocessing steps (e.g., global z-score normalization) consistently degraded FLENI performance. These insights emphasize the importance of preprocessing consistency and spatial completeness in training models intended for clinical generalization.

Saliency analyses provided additional confirmation of these findings, showing stronger attention to regions in ADNI, but attenuated or shifted attention in FLENI. These differences likely reflect both demographic variability and acquisition protocol discrepancies between cohorts.

\section{Conclusion}

This work presents a comprehensive evaluation of deep learning models for Alzheimer’s disease diagnosis using FDG-PET scans, with a focus on validating generalization across geographic and demographic boundaries. Our results show that all evaluated architectures—Inception, Transformer, and P-ResNet—performed comparably, achieving high AUC scores on ADNI.

However, when tested on the out-of-distribution FLENI datasets, every model exhibited at least a 10-point drop in AUC, despite using optimized configurations. This consistent drop highlights the risks of relying solely on in-domain validation and underscores the necessity of cross-cohort benchmarking when deploying diagnostic models in real-world clinical environments. 

Future work should investigate training with more diverse global datasets, as well as domain adaptation and harmonization techniques to reduce performance degradation and improve fairness across underrepresented populations.

%% file: main.bbl
\begin{thebibliography}{13}
\providecommand{\natexlab}[1]{#1}
\providecommand{\url}[1]{\texttt{#1}}
\expandafter\ifx\csname urlstyle\endcsname\relax
  \providecommand{\doi}[1]{doi: #1}\else
  \providecommand{\doi}{doi: \begingroup \urlstyle{rm}\Url}\fi

\bibitem[Arora et~al.(2023)Arora, Alderman, Palmer, Ganapathi, Laws, McCradden,
  Oakden-Rayner, Pfohl, Ghassemi, McKay, Treanor, Rostamzadeh, Mateen, Gath,
  Adebajo, Kuku, Matin, Heller, Sapey, Sebire, Cole-Lewis, Calvert, Denniston,
  and Liu]{Arora2023}
Arora, A., Alderman, J.~E., Palmer, J., Ganapathi, S., Laws, E., McCradden,
  M.~D., Oakden-Rayner, L., Pfohl, S.~R., Ghassemi, M., McKay, F., Treanor, D.,
  Rostamzadeh, N., Mateen, B., Gath, J., Adebajo, A.~O., Kuku, S., Matin, R.,
  Heller, K., Sapey, E., Sebire, N.~J., Cole-Lewis, H., Calvert, M., Denniston,
  A., and Liu, X.
\newblock The value of standards for health datasets in artificial
  intelligence-based applications.
\newblock \emph{Nature Medicine}, 29\penalty0 (11):\penalty0 2929–2938,
  October 2023.
\newblock ISSN 1546-170X.
\newblock \doi{10.1038/s41591-023-02608-w}.
\newblock URL \url{http://dx.doi.org/10.1038/s41591-023-02608-w}.

\bibitem[Ch{\'e}telat et~al.(2020)Ch{\'e}telat, Arbizu, Barthel, Garibotto,
  Law, Morbelli, van~de Giessen, Agosta, Barkhof, Brooks,
  et~al.]{chetelat2020amyloid}
Ch{\'e}telat, G., Arbizu, J., Barthel, H., Garibotto, V., Law, I., Morbelli,
  S., van~de Giessen, E., Agosta, F., Barkhof, F., Brooks, D.~J., et~al.
\newblock Amyloid-pet and 18f-fdg-pet in the diagnostic investigation of
  alzheimer's disease and other dementias.
\newblock \emph{The Lancet Neurology}, 19\penalty0 (11):\penalty0 951--962,
  2020.

\bibitem[Ding et~al.(2019)Ding, Sohn, Kawczynski, Trivedi, Harnish, Jenkins,
  Lituiev, Copeland, Aboian, Mari~Aparici, et~al.]{ding2019deep}
Ding, Y., Sohn, J.~H., Kawczynski, M.~G., Trivedi, H., Harnish, R., Jenkins,
  N.~W., Lituiev, D., Copeland, T.~P., Aboian, M.~S., Mari~Aparici, C., et~al.
\newblock A deep learning model to predict a diagnosis of alzheimer disease by
  using 18f-fdg pet of the brain.
\newblock \emph{Radiology}, 290\penalty0 (2):\penalty0 456--464, 2019.

\bibitem[Frisoni et~al.(2010)Frisoni, Fox, Jack, Scheltens, and
  Thompson]{frisoni2010clinical}
Frisoni, G.~B., Fox, N.~C., Jack, C.~R., Scheltens, P., and Thompson, P.~M.
\newblock The clinical use of structural mri in alzheimer disease.
\newblock \emph{Nature Reviews Neurology}, 6\penalty0 (2):\penalty0 67–77,
  February 2010.
\newblock ISSN 1759-4766.
\newblock \doi{10.1038/nrneurol.2009.215}.
\newblock URL \url{http://dx.doi.org/10.1038/nrneurol.2009.215}.

\bibitem[He et~al.(2016)He, Zhang, Ren, and Sun]{resnet}
He, K., Zhang, X., Ren, S., and Sun, J.
\newblock Deep residual learning for image recognition.
\newblock In \emph{2016 IEEE Conference on Computer Vision and Pattern
  Recognition (CVPR)}, pp.\  770--778, 2016.
\newblock \doi{10.1109/CVPR.2016.90}.

\bibitem[He et~al.(2023)He, Gan, Li, Rekik, Yin, Ji, Gao, Wang, Zhang, and
  Shen]{he2023transformers}
He, K., Gan, C., Li, Z., Rekik, I., Yin, Z., Ji, W., Gao, Y., Wang, Q., Zhang,
  J., and Shen, D.
\newblock Transformers in medical image analysis.
\newblock \emph{Intelligent Medicine}, 3\penalty0 (1):\penalty0 59--78, 2023.

\bibitem[Jun et~al.(2024)Jun, Jeong, Heo, and Suk]{Jun2024}
Jun, E., Jeong, S., Heo, D.-W., and Suk, H.-I.
\newblock Medical transformer: Universal encoder for 3-d brain mri analysis.
\newblock \emph{IEEE Transactions on Neural Networks and Learning Systems},
  35\penalty0 (12):\penalty0 17779–17789, December 2024.
\newblock ISSN 2162-2388.
\newblock \doi{10.1109/tnnls.2023.3308712}.
\newblock URL \url{http://dx.doi.org/10.1109/TNNLS.2023.3308712}.

\bibitem[Li et~al.(2023)Li, Chen, Tang, Wang, Landman, and
  Zhou]{li2023transforming}
Li, J., Chen, J., Tang, Y., Wang, C., Landman, B.~A., and Zhou, S.~K.
\newblock Transforming medical imaging with transformers? a comparative review
  of key properties, current progresses, and future perspectives.
\newblock \emph{Medical image analysis}, 85:\penalty0 102762, 2023.

\bibitem[Shamshad et~al.(2023)Shamshad, Khan, Zamir, Khan, Hayat, Khan, and
  Fu]{shamshad2023transformers}
Shamshad, F., Khan, S., Zamir, S.~W., Khan, M.~H., Hayat, M., Khan, F.~S., and
  Fu, H.
\newblock Transformers in medical imaging: A survey.
\newblock \emph{Medical image analysis}, 88:\penalty0 102802, 2023.

\bibitem[Vaswani et~al.(2017)Vaswani, Shazeer, Parmar, Uszkoreit, Jones, Gomez,
  Kaiser, and Polosukhin]{vaswani2017attention}
Vaswani, A., Shazeer, N., Parmar, N., Uszkoreit, J., Jones, L., Gomez, A.~N.,
  Kaiser, {\L}., and Polosukhin, I.
\newblock Attention is all you need.
\newblock \emph{Advances in neural information processing systems}, 30, 2017.

\bibitem[Weiner et~al.(2015)Weiner, Veitch, Aisen, Beckett, Cairns, Green,
  Harvey, Jack, Jagust, Liu, et~al.]{weiner2015impact}
Weiner, M.~W., Veitch, D.~P., Aisen, P.~S., Beckett, L.~A., Cairns, N.~J.,
  Green, R.~C., Harvey, D., Jack, C.~R., Jagust, W., Liu, E., et~al.
\newblock Impact of the alzheimer's disease neuroimaging initiative, 2004 to
  2014.
\newblock \emph{Alzheimer's \& Dementia}, 11\penalty0 (7):\penalty0 865--884,
  2015.

\bibitem[{World Health Organization}(2021)]{WHO2021dementia}
{World Health Organization}.
\newblock Dementia, 2021.
\newblock URL \url{https://www.who.int/news-room/fact-sheets/detail/dementia}.

\bibitem[Yang et~al.(2024)Yang, Zhang, Gichoya, Katabi, and Ghassemi]{Yang2024}
Yang, Y., Zhang, H., Gichoya, J.~W., Katabi, D., and Ghassemi, M.
\newblock The limits of fair medical imaging ai in real-world generalization.
\newblock \emph{Nature Medicine}, 30\penalty0 (10):\penalty0 2838–2848, June
  2024.
\newblock ISSN 1546-170X.
\newblock \doi{10.1038/s41591-024-03113-4}.
\newblock URL \url{http://dx.doi.org/10.1038/s41591-024-03113-4}.

\end{thebibliography}
